\documentclass{article}

\usepackage{PRIMEarxiv}

\usepackage[utf8]{inputenc} 
\usepackage[T1]{fontenc}    
\usepackage{hyperref}       
\usepackage{url}            
\usepackage{booktabs}       
\usepackage{amsfonts}       
\usepackage{nicefrac}       
\usepackage{microtype}      
\usepackage{lipsum}
\usepackage{fancyhdr}       
\usepackage{graphicx}       
\usepackage{cite}
\usepackage{amsmath,amssymb,amsfonts}
\usepackage{algorithmic}
\usepackage{graphicx}
\usepackage{textcomp}
\usepackage{booktabs}
\usepackage{diagbox}
\usepackage[justification=centering]{caption}
\graphicspath{{media/}}     

\title{Using frequency attention to make adversarial patch powerful against person detector
\thanks{
The first author and the second author have the same contribution to the work. 
} 
}

\author{
 Xiaochun Lei \\
  School of Computer Science and Information Security\\
  Guilin University of Electronic Technology\\
  China, GuiLin 541010 \\
  \texttt{lxc8125@guet.edu.cn} \\
   \And
 Chang Lu \\
  School of Computer Science and Information Security\\
  Guilin University of Electronic Technology\\
  China, GuiLin 541010 \\
  \texttt{Changlu@keter.top} \\
  \And
  Zetao Jiang$^*$\\
  School of Computer Science and Information Security\\
  Guilin University of Electronic Technology\\
  China, GuiLin 541010 \\
  \texttt{zetaojiang@guet.edu.cn} \\
  \And
  Zhaoting Gong \\
  School of Computer Science and Information Security\\
  Guilin University of Electronic Technology\\
  China, GuiLin 541010 \\
  \texttt{gavin@gong.host} \\
  \And
  Xiang Cai \\
  School of Computer Science and Information Security\\
  Guilin University of Electronic Technology\\
  China, GuiLin 541010 \\
  \texttt{xiangcai@keter.top} \\
  \And
  Linjun Lu \\
  School of Computer Science and Information Security\\
  Guilin University of Electronic Technology\\
  China, GuiLin 541010 \\
  \texttt{linjunlu@zerorains.top} \\
  \And
}

\begin{document}
\maketitle

\begin{abstract}
Deep neural networks (DNNs) are vulnerable to adversarial attacks. 
In particular, object detectors may be attacked by applying a particular adversarial patch to the image.
However, because the patch shrinks during preprocessing, most existing approaches that employ adversarial patches to attack object detectors would diminish the attack success rate on small and medium targets.
This paper proposes a Frequency Module(FRAN), a frequency-domain attention module for guiding patch generation.
This is the first study to introduce frequency domain attention to optimize the attack capabilities of adversarial patches.
Our method increases the attack success rates of small and medium targets by 4.18\% and 3.89\%, respectively, over the state-of-the-art attack method for fooling the human detector while assaulting YOLOv3 without reducing the attack success rate of big targets.

\end{abstract}


\section{Introduction}

Deep neural networks (DNNs) have garnered much success in computer vision.
DNNs have proven to be quite effective in data-driven computer vision tasks and multiple areas such as autopilot, aviation, and medicine.
However, DNNs have shown flaws susceptible to perturbation by adversarial samples in image classification, object tracking, and object detection.
Previous researches\cite{IEEEexample:Goodfellow2015ExplainingAH,IEEEexample:Kurakin2017AdversarialML,IEEEexample:Phan2021AdversarialIP} explored attack classification networks. 
In the last years, several studies\cite{IEEEexample:Thys2019FoolingAS,IEEEexample:Liu2018DPatchAO,IEEEexample:Wu2020MakingAI,IEEEexample:Xu2020AdversarialTE} have recently demonstrated that object detection networks are similarly susceptible to perturbation by adversarial samples. 
Thys et al.\cite{IEEEexample:Thys2019FoolingAS} create a printed adversarial patch that successfully attacked the object detector YOLOv2\cite{IEEEexample:Redmon2017YOLO9000BF}.
Liu et al.\cite{IEEEexample:Liu2018DPatchAO} successfully render the object detection network incapable of recognizing objects and generating incorrect prediction categories. 
Wu et al.\cite{IEEEexample:Wu2020MakingAI} formed wearable clothes for eluding the object detector. 
They use one adversarial patch to attack one-stage\cite{IEEEexample:Redmon2018YOLOv3} and two-stage\cite{IEEEexample:Ren2015FasterRT} object detectors.
Wu et al.\cite{IEEEexample:Wu2020MakingAI} create wearable clothes for evading the object detector.
Xu et al.\cite{IEEEexample:Xu2020AdversarialTE} use a thin-plate spline (TPS) base transformer to build a more robust patch that can withstand non-rigid deformation caused by a moving person's posture changes.

Furthermore, network input size, JPEG compression\cite{IEEEexample:Goodfellow2015ExplainingAH}, and network processing can all have impacted the efficacy of adversarial patches. 
When the picture is scaled, the loss of high-frequency signals significantly decreases faster than the loss of low-frequency signals. 
The reduction in picture size resulted in an overall loss of high-frequency signals, leading to imbalanced representations. 
High-frequency information is also discarded by JPEG compression.
As described, using patches to attack the object detector has an inherent difficulty: while patching, we frequently must shrink the patches and then paste them on the objects in the image. 
The patch's attack ability is reduced due to the reduction in pixels caused by sampling. 

This article built the Frequency Module(FRAN), a frequency attention module, to guide the patch generation process.
This module can determine the relative importance of various locations within the patch frequency domain, allowing the patched attack to rely on low-frequency signals. 
The attack success rates of medium and small objects in our method are 3.89\% and 4.18\%, respectively, higher than the current best attack algorithm\cite{IEEEexample:Thys2019FoolingAS}, which uses the patch to fool YOLOv3 while detecting people without affecting the attack success rate of large objects.

Our contributions are summarized as follows:
\begin{itemize}
\item[1.] We used a frequency attention module to guide the patch generation process in the frequency domain, increasing the adversarial patch's attack success rate against small and medium objects.
\item[2.] Our method raises the patch's robustness when defending against adversarial attacks using JPEG compression technique.
\item[3.] We proposed three new indicators, ASRm, ASRl, and ASRs, to measure the patch's attack effect on objects of various sizes.
\end{itemize}

The rest of this paper is structured as follows: 
Section \ref{Related} goes over the related work on adversarial attacks. 
Section \ref{Generate} discusses how we generate patches and how we use the frequency attention module to guide the process of patch generation.
In Section \ref{Experiments}, we test the efficacy of the patch.
In Section \ref{Conclusion and Future Work}, we reach a conclusion and look at the future directions of our research.

\section{Related work}  

\label{Related}

Attacks against object detectors have grown exceedingly prevalent due to the widespread use of object detection networks in various areas. 
In this section, we first go over the related works of the object detection network and the attacks against the object detection network, and then we go over the present challenges.

\subsection{Object detection}

The popular one-stage object detector and two-stage object detector are the most common types of modern detectors.
The two-stage detector splits the detection task into two parts. 
In the first stage, the objects in the picture are located. 
And then, the object detector outputs the categories of the located objects acquired in the first stage.
Faster RCNN \cite{IEEEexample:Ren2015FasterRT} is a common two-stage detection network. 
It extracts the region of interest (ROI) using RPN, then produces and classifies detection boxes based on the ROI.
Although two-stage object detectors have the advantage of accuracy, the detection speed of two-stage object detectors is often slower than that of one-stage object detectors since the prediction process is divided into two stages.

The one-stage detector uses the entire picture as input and outputs the position and category of the detection boxes directly in the output layer.
Yolo-Series algorithms \cite{IEEEexample:Redmon2017YOLO9000BF,IEEEexample:Redmon2016YouOL,IEEEexample:Redmon2018YOLOv3} are exceptionally representative one-stage object detectors. 
To acquire detection boxes, they do not need to extract ROI. 
Yolo-Series methods have been the most often used object detectors due to their quick detection speed and good detection effect.
In this paper, we target YOLOv3 \cite{IEEEexample:Redmon2018YOLOv3} as the main victim detector.

\subsection{Adversarial attacks on object detector}

According to the methods of adversarial attack, the job of attacking the object detector can be separated into two groups. 
One group is that the object detector can be attacked by rendering an adversarial patch over all of the objects in the picture. 
The other is to attack the detectors by changing the pixel value of the input image directly.

Thys et al.\cite{IEEEexample:Thys2019FoolingAS} first proposed using the adversarial patch to attack the object detection network in 2019.
They use the loss function to guide the patch generation process, gradient descent to update the patch, and eventually, disable YOLOv2\cite{IEEEexample:Redmon2017YOLO9000BF} from detecting patched objects.
Moreover, Thys et al. print out the patches to fool the object detector in the physical world by holding a small cardboard plate in front of their body.
Liu et al.\cite{IEEEexample:Liu2018DPatchAO} interfere with the detection results of the entire picture by affixing patches to the image's edge angle and implemented the object-vanishing and object-mislabeling attacks by designing different loss functions in the same year.
Wu et al.\cite{IEEEexample:Wu2020MakingAI} use one adversarial patch to attack one-stage\cite{IEEEexample:Redmon2017YOLO9000BF,IEEEexample:Redmon2018YOLOv3} and two-stage\cite{IEEEexample:Ren2015FasterRT} object detectors. 
They design wearable clothes that could attack the object detector in the physical world by printing the patch on clothes.
Furthermore, they explore how patches' attack ability can be applied among different detectors.
According to their research, the patch's attack has strong transferability between networks with the same backbone but is poor between one-stage and two-stage object detection networks.
Nevertheless, there are several issues when using adversarial patches to attack surveillance systems in the physical world. 
One issue is that the object detector may observe the adversarial patches from different viewpoints in the physical world. 
There is much work\cite{IEEEexample:Xu2020AdversarialTE,IEEEexample:Bookstein1989PrincipalWT,IEEEexample:Lennon2021PatchAI} to be done in order for the patch to be sufficiently robust from several viewpoints.
Xu et al.\cite{IEEEexample:Xu2020AdversarialTE} utilize a TPS base transformer to model the non-rigid deformation of clothing induced by a moving person's posture changes to create a more robust adversarial T-shirt.
Lennon et al.\cite{IEEEexample:Lennon2021PatchAI} apply an affine transformation to generate distinct viewpoints of the same patch, fuse the results of computation generated by different viewpoints patches, and then update the patch using gradient descent. 
As a result, the created patch is more robust from various viewpoints.

In previous research, people directly modified the picture pixels to attack the object detection network.
This attack method originates from the study \cite{IEEEexample:Goodfellow2015ExplainingAH} that concerned attack classification networks, and its primary purpose is to create a disturbance that is difficult to notice by humans and causes the object detector to be disturbed at the same time.
DAG\cite{IEEEexample:Xie2017AdversarialEF} is proposed in Xie et al.'s work to attack the object detection network.
It focuses on attacking Faster RCNN\cite{IEEEexample:Ren2015FasterRT}.
They first give each object an adversarial label selected randomly from other incorrect classes.
In the next step, the image will be updated based on the detector's output.
There was also some work\cite{IEEEexample:Wei2019TransferableAA} against the transferability of this attack method across networks.
According to the latest research, \cite{IEEEexample:Chow2020TOGTA} can attack one-stage target detection networks and two-stages as well by changing the picture's pixels.
Their work describes three Targeted Adversarial Objectness Gradient (TOG) attack methods that can induce object-vanishing, object-fabrication, and object-mislabeling attacks in state-of-the-art object detectors.

There is an inherent difficulty with the strategy of using patches to attack the object detector: while patching, we frequently need to shrink the patches and then paste them on the objects in the image. 
The reduction in pixels caused by sampling will diminish the attack ability of the patches.
As a result, the effect of current attack approaches using the adversarial patch for small and medium target objects needs to be improved.
We guide the frequency domain information of the patch through the frequency attention module, such that the patch's attack ability depends more on the low-frequency signal.
Because the low-frequency signal of the picture will lose less than the high-frequency signal during image shrinking, the patch's aggressiveness will be preserved even more.

\begin{figure*}[t]
	\centering 
	\includegraphics[width=1.0\textwidth]{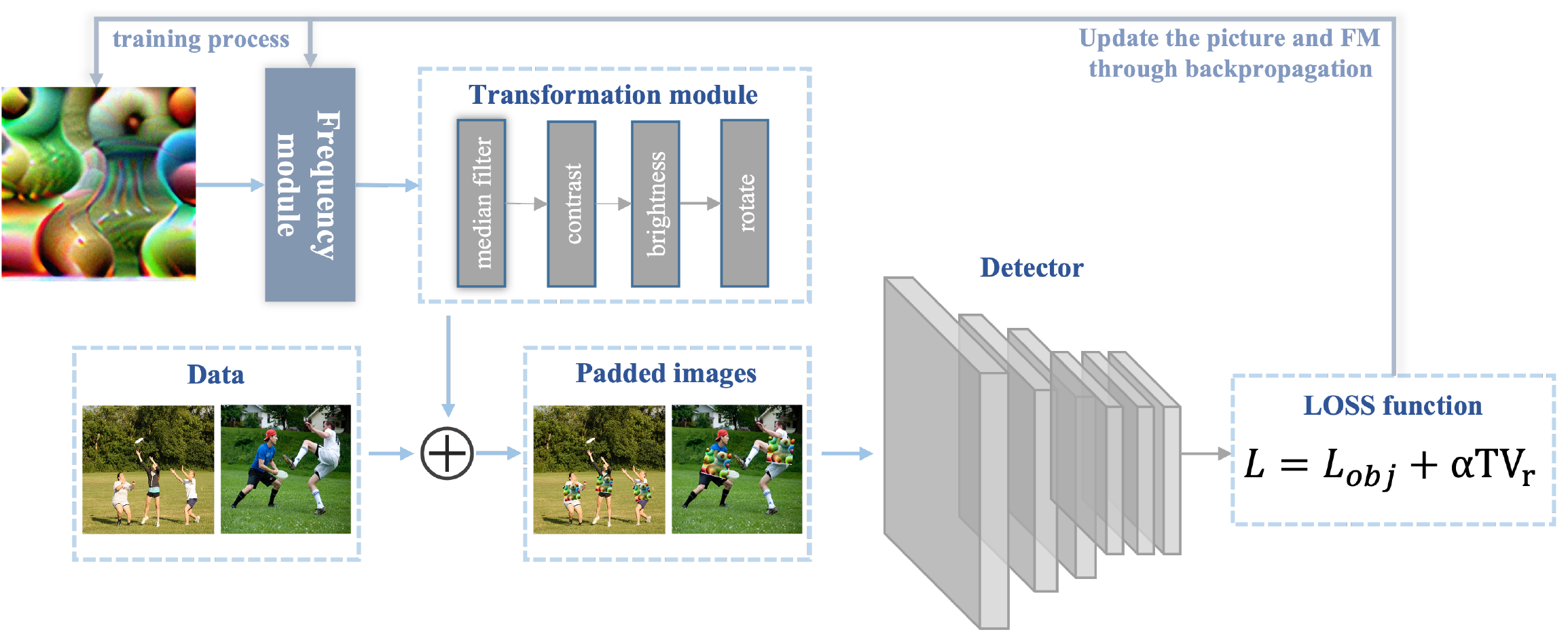} 
	\caption{A overview of the framework.} 
	\label{fig.Overview} 
\end{figure*}

\begin{figure*}[t]
	\centering 
	\includegraphics[width=0.8\textwidth]{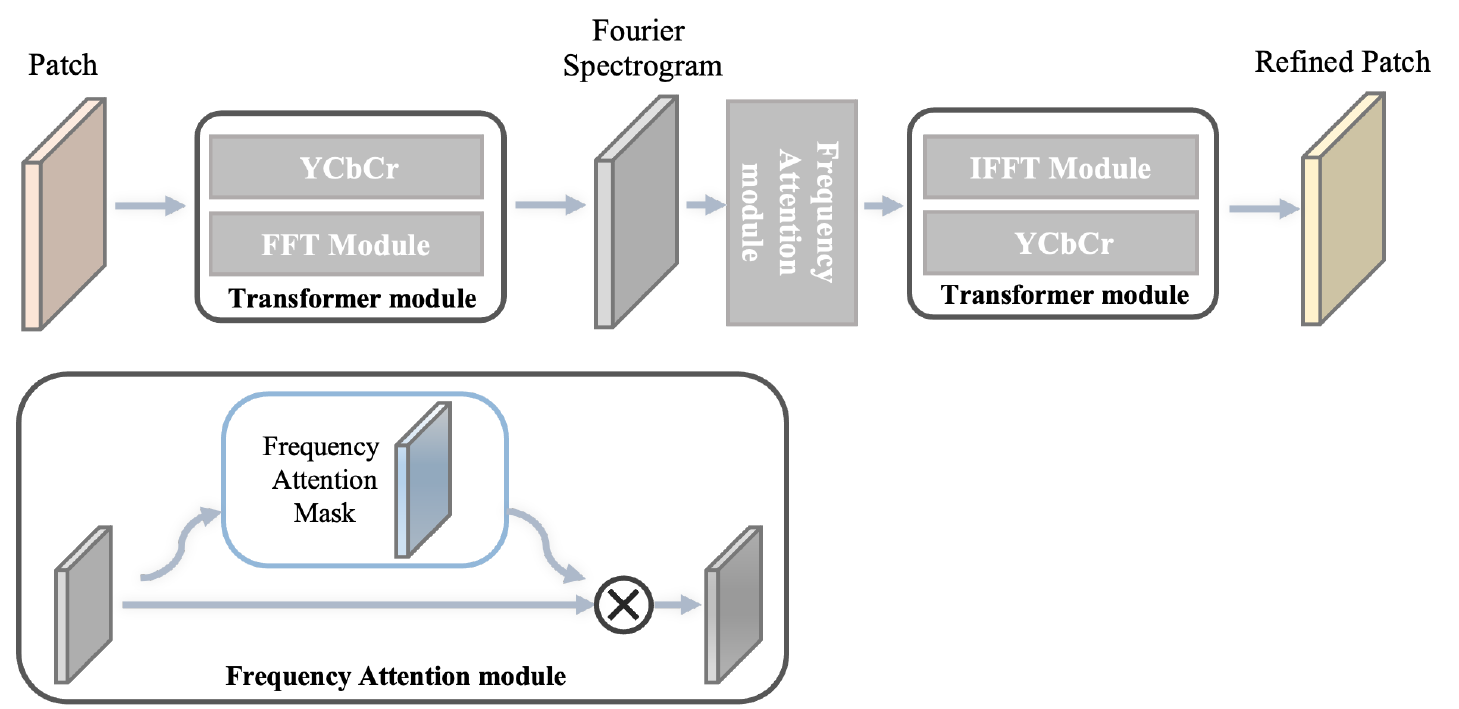} 
	\caption{
		Frequency module. 
		First, the patch is converted into YCbCr color space. 
		Then, the patch is sent into FFT module to be converted into a spectrum diagram. 
		After that, the spectrum diagram multiplies the frequency attention mask.
		The patch is then delivered to IFFT module, translating from YCbCr to RGB color space to generate FRAN's output.
	} 
	\label{fig.Frequency} 
\end{figure*}

\section{Generate more robust adversarial patch}

\label{Generate}

The current section describes in detail our method for creating adversarial patches. 
Based on the problems outlined in Section \ref{Related}, we present a solution that employs a frequency attention module to guide the patch generating process. 
Our method directs the learning of the patch's frequency information by FRAN.
Figure \ref{fig.Overview} illustrates the primary process of generating the patche $\mathcal{P} \in \mathbb{R}^{w \times h \times 3}$.
In Figure \ref{fig.Overview}, Detector is the victim detector\cite{IEEEexample:Wu2020MakingAI}.
Frequency Module(FRAN) is the frequency domain attention module and Data is the training data set.
We first add the patch and FRAN parameters to the optimizer and freeze the detector network parameters before training.
We solely use gradient descent to update the patch's pixel value and FRAN's parameters during the training procedure.
In the training procedure, the patches begin by walking through FRAN with the given data set.
The transformation module will render the rescaled patch over every person from the ground truth in the input image.
The image is then fed into the victim detector, and the loss function value is calculated based on the detection results.
Finally, we update the patch's pixel value and the FRAN's parameters based on the calculation result of the loss function. 
Our ultimate goal is to create patches that minimize the detection success rate of pedestrians of various sizes throughout a dataset.

\subsection{Frequency module}

\label{fm}

FRAN is a frequency domain attention module that assigns different weights in the frequency domain.
Figure \ref{fig.Frequency} depicts the construction of FRAN. 
The FRAN's general procedure entails converting the patch to the frequency domain using Fast Fourier Transform (FFT)\cite{IEEEexample:Cooley1965AnAF} and multiplying it by the attention mask.
The output of FRAN is obtained by applying an Inverse Fast Fourier Transform (IFFT) to the results obtained in the previous step.

More formally, we take $\mathcal{F}_{\theta }$ as an attention function in the frequency domain.
The attention function takes the patch $\mathcal{P} \in \mathbb{R}^{w \times h \times 3}$ as input and $\theta \in \mathbb{R}^{w \times h \times 3}$ as the parameter of the attention module.
$FFT(x)$ is the fast fourier transform while $FFT^{-1}(x)$ represents the inverse fast fourier transform.
$\mathcal{Y}(x)$ transforms the patch from RGB to YCbCr color format.
$\mathcal{Y}^{-1}(x)$ is the inverse process of $\mathcal{Y}(x)$ 
FRAN's output $\mathcal{F}_{\theta }(\mathcal{P})$ can be stated below:

\begin{equation}
\mathcal{F}_{\theta }(\mathcal{P}) = \mathcal{Y}^{-1}(FFT^{-1}(FFT(\mathcal{Y}(\mathcal{P}\odot \theta))))
\end{equation}

Because the image's high-frequency signal generally loss more than the low-frequency signal during scaling.
Consequently, the attack success rate for small and medium target objects can be enhanced if the patch's attack ability depends more on the low-frequency signal.

\subsection{TV loss}

\label{tv_loss}

TV loss\cite{IEEEexample:Thys2019FoolingAS} represents the total variation in the image, resulting in a patch with smooth color transitions and minimal noise. 
The following formula shows how TV loss is calculated in the previous works\cite{IEEEexample:Thys2019FoolingAS,IEEEexample:Wu2020MakingAI,IEEEexample:Xu2020AdversarialTE} of attacking object detectors with adversarial patches.

\begin{equation}
L_{t v}=\sum_{i, j} \sqrt{\left(\left(p_{i, j}-p_{i+1, j}\right)^{2}+\left(p_{i, j}-p_{i, j+1}\right)^{2}\right.}
\end{equation}

The $p_{i,j}$ represents the pixel in patch $\mathcal{P}$.
This approach only calculates the distance between each pixel and the pixel to its right and the pixel below it. 
To better smooth the resulting patch, we compute the distance between each pixel and the eight pixels. Thus, we calculate TV loss $TVr$ as following:

\begin{equation} 
\label{equ.TVr}
TV_r = \sum_{i,j}  \sqrt{\sum_{k \in \{-1,0,1\}} \sum_{m \in \{-1,0,1\}} (p_{i,j}-p_{i+k,j+m})^2}
\end{equation}

$TV_r$ can better measure the smoothness of a picture since it measures the distance of surrounding pixels among each pixel.

\subsection{Loss function}

Our optimization goal is divided into two parts: 

\begin{itemize}
\item $TV_r$: The image's total variance; this loss guarantees that the image is smooth and does not cause excessive noise. For further information, see Section \ref{tv_loss}.
\item $L_{obj}$: The highest objectness score from the prediction.
As previously mentioned \cite{IEEEexample:Thys2019FoolingAS,IEEEexample:Wu2020MakingAI}, this loss can lower the detector output's objectness score while detecting the person. 
Set $\boldsymbol{x}$ to be the input image and $\mathcal{P}$ to be the patch. 
The transformation function $\mathcal{T}(\boldsymbol{x},\mathcal{P})$ accepts photos and patches as input and scales and affixes the patches to the person in the image.
The frequency attention function $\mathcal{F}_\theta$ was discussed in Section \ref{fm}.
A detector takes a patched image $\mathcal{T}(\boldsymbol{x},\mathcal{F}_\theta(\mathcal{P}))$ as its input and outputs the result with objectness scores $\mathcal{D}_i(\mathcal{T}(\boldsymbol{x},\mathcal{F}_\theta(\mathcal{P})))$ is defined as follows:

\begin{equation}
	L_{obj} = \sum_i \mathcal{D}_i(\mathcal{T}(\boldsymbol{x},\mathcal{F}_\theta(\mathcal{P})))
\end{equation}

\end{itemize}

Our total loss function is made up of these two parts:

\begin{equation}
L = L_{obj} + \alpha TV_r
\end{equation}

We combine the two losses, scaled by empirically established factors $alpha$ and use Adam\cite{IEEEexample:Kingma2015AdamAM} to optimize the loss.

\section{Experiments and results}  

\label{Experiments}
\begin{figure*}[t]
	\centering 
	\includegraphics[width=0.7\textwidth]{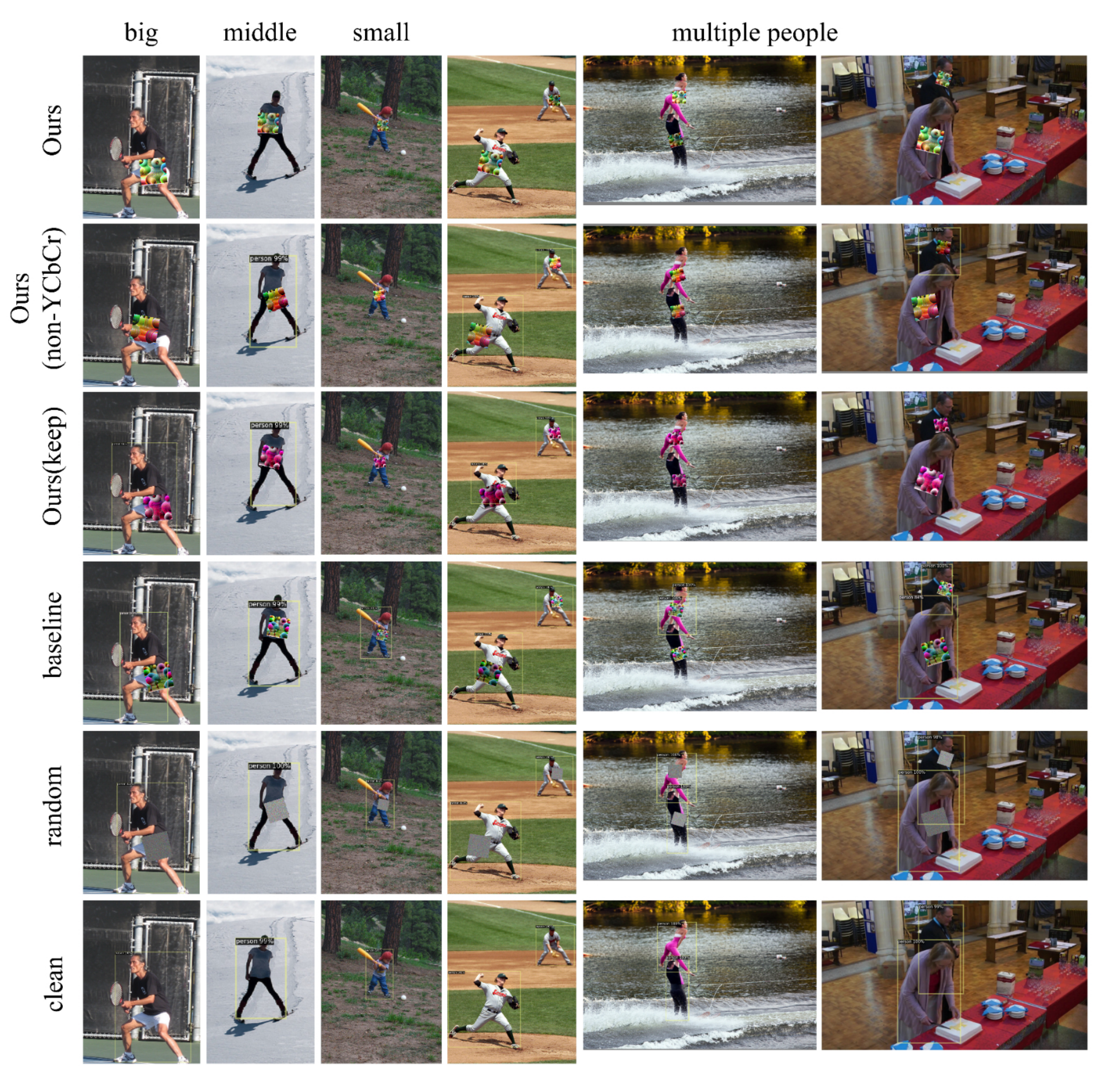} 
	\caption{The effectiveness of an adversarial patch in evading YOLOv3 detection. 
	We classify the images from the test set into four groups based on the size and number of people.
	Except for the last rows, each row employs a distinct attack method.
	First row: use FRAN to direct the generation of adversarial patches in the first seven epochs.
	Secend row: use FRAN to direct the generation of adversarial patches in the first seven epochs but did not convert the patch to YCbCr color space.
	Third row: use FRAN throughout the generating process.
	Fourth row: use the baseline\cite{IEEEexample:Thys2019FoolingAS} method to generate the patch.
	Fifth row: use random generated patch.
	Sixth row: original image without patch} 
	\label{fig.total} 
\end{figure*}

We demonstrate the efficiency of our strategy in this section by comparing our attack effect against several object detectors.
While there are many researches that have optimized the aggressiveness of the patch, their research has focused on optimizing the aggressiveness of the patch in the physical world\cite{IEEEexample:Wu2020MakingAI,IEEEexample:Xu2020AdversarialTE,IEEEexample:Lennon2021PatchAI} or generating new attack effects\cite{IEEEexample:Liu2018DPatchAO,IEEEexample:Chow2020TOGTA}.
The core idea to generate the adversarial patch of those researches is still based on the baseline\cite{IEEEexample:Thys2019FoolingAS}.
Thus, we choose the method proposed in \cite{IEEEexample:Thys2019FoolingAS} as the baseline.
We refer to the patch in our methods as $\mathcal{P}_1$ and the patch generated by the baseline as $\mathcal{P}_2$.

The rest of this section is structured as following.
In Section \ref{exset}, we introduce the basic settings of the experiment, including the data set, the attacked object detectors and the evaluation index.
Section \ref{result} introduces our method for attacking objects of various sizes and different detectors then evaluates its robustness against JPEG compression resistance.
Finally, we use ablation experiments to illustrate the efficiency of our strategy.
Figure \ref{fig.total} depicts several examples of attacking the object detector.
We compared four attack methods in the figure.
The first three methods employ the FRAN to guide the patch generation process, and their impacts on small and medium objects are more significant than the Baseline.
In the situation of object overlap in a multi-person scene, our method can also have a high attack success rate.
The last two lines of the image show the results of the object detector when detecting the original image and the image with a random patch. 
People in those images can be successfully detected.

\subsection{Experimental setup}
\label{exset}

\subsubsection{Evaluation metrics}

\label{section.asr}

This section will introduce the evaluation metrics ASR\cite{IEEEexample:2020Understanding}, ASRs, ASRm, and ASRl to analyze adversarial patch attacks.

The attack success rate (ASR) \cite{IEEEexample:2020Understanding} is a metric to better evaluate the attack object detector's performance.
\cite{IEEEexample:2020Understanding} proposes three different types of ASRs. 
The ASR is used in this work to counter the object-vanishing attack, which is defined as following:

\begin{equation}\label{model3_coef}
        ASR=\frac{\sum_{\boldsymbol{x} \in \mathcal{D}} \sum_{\hat{\boldsymbol{o}} \in \hat{\mathcal{O}}(\boldsymbol{x})} \mathbb{1}\left[\neg \exists \hat{\boldsymbol{o}}^{\prime} \in \hat{\mathcal{O}}\left(\boldsymbol{x}^{\prime}\right)\left(\operatorname{IOU}\left(\hat{\boldsymbol{o}}, \hat{\boldsymbol{o}}^{\prime}\right) \leq t_{\mathrm{IOU}}\right)\right]}{\sum_{\boldsymbol{x} \in \mathcal{D}}\|\hat{\mathcal{O}}(\boldsymbol{x})\|}
\end{equation}

If $\mathbb{1}[condition]$ is 1 or 0, the condition is satisfied.
We consider $\boldsymbol{x}$ as the input image, $\mathcal{D}$ as the set of all $\boldsymbol{x}$.
$\boldsymbol{x'}$ represents the image after adversarial disturbance.
We use $\hat{\boldsymbol{o}}$ to denote a ground truth box and $\hat{\boldsymbol{o'}}$ to denote the predicted box.
Let $\hat{\boldsymbol{O}}{\boldsymbol{(x)}}$ be the set of ground-truth objects of $\hat{\boldsymbol{o}}$. 
Similarly, let $\hat{\boldsymbol{O}}{\boldsymbol{(x')}}$ be the set of predicted objects of $\hat{\boldsymbol{o'}}$. 
$IOU(\hat{\boldsymbol{o}}, \hat{\boldsymbol{o}}^{\prime})$ represents the intersection over union of $\boldsymbol{o}$ and $\boldsymbol{o'}$.
$t_{IOU}$, a specified super parameter, represents the IOU threshold. 
For the calculations in the experiment, we choose 0.5\cite{IEEEexample:2020Understanding}.

However, calculating ASR alone does not account for the attack effect of adversarial patches on objects of numerous sizes; consequently, we proposed ASRs, ASRm, and ASRl based on the ideas of APs, APm, and APl \cite{IEEEexample:2014Microsoft}. 
ASRs, ASRm, and ASRl represent the attack success rates of attacks on small, medium, and large objects, respectively.

We utilized height to differentiate between three distinct item sizes.
Let $\mathcal{H}$ stand for a height function, $\mathcal{H}(\hat{\boldsymbol{o}})$ for the detection box's height, and $\mathcal{H}(\boldsymbol{x})$ for a picture's height. 
Next, the height ratio $\epsilon$ can be defined as follows:

\begin{equation}
\epsilon = \frac{\mathcal{H}(\hat{\boldsymbol{o})}}{\mathcal{H}(\boldsymbol{x})}
\end{equation}

We made two boundaries $e_0$ and $e_1$ to separate detection boxes of different sizes. 
Small objects have an $e$ value less than $e_0$, while medium objects have an $e$e value between $e_0$ and $e_1$.
A big object is one with an $e$e larger than $e_1$. 
In our research, $\epsilon_0$ is 0.3,and $\epsilon_1$ is 0.6.
While dividing the data set, this two values can balance the number of three sizes of objects.

\subsubsection{Data collection}

We used two datasets when training the object detector: COCO\cite{IEEEexample:2014Microsoft} and VOC\cite{IEEEexample:Everingham2014ThePV}. 
Similarly, we chose 16000 images from COCO dataset to generate the adversarial patch. 
After randomly shuffling the dataset, we used 12000 images as the training dataset and 4000 images as the test dataset.
There is at least one person in each of the images. 
The number of the small, medium and big size of human boxes is nearly the same in the training and test datasets. 
We utilized ASR, ASRs, ASRM and ASRl described in Section \ref{section.asr} as the evaluation metrics of the attack effectiveness.

\subsubsection{Victim object detectors}

\label{section.set}

Our work selected YOLOv3\cite{IEEEexample:Redmon2018YOLOv3}, YOLOv4\cite{IEEEexample:YOLOv4} and RetinaNet\cite{IEEEexample:RetinaNet} as the victim detectors, and the input image size is $3 \times 612 \times 612$.
We first trained pre-training models on the COCO dataset and then trained the final models using the VOC dataset.

\subsubsection{Implementation details}

We randomly initialized a $3 \times 950 \times 950$ patch in the patch generation process.
We used FRAN to guide patch production in the first seven training epochs. Moreover, Adam\cite{IEEEexample:Kingma2015AdamAM} was used in our experiments. 
For training, we used a single Tesla A100. 
Our batch size in the training process was four, and the learning rate was set at 0.005.
Each training stoped after about 45 hours when ASR rarely rises. 
We used the patch with the highest ASR in the test dataset as the final training result.

\subsection{Results}

\label{result}

Our method achieved 93.68\% ASR when attacking YOLOv3\cite{IEEEexample:Redmon2018YOLOv3}.
Simultaneously, our method successfully raised ASRm and ASRs without affecting ASRl.
Our ASR improved by 2.56\%, ASRm increased by 3.89\%, and ASRl amplified by 4.18\% compared with the baseline\cite{IEEEexample:Thys2019FoolingAS}.

\subsubsection{Attack on different size objects}

\begin{figure}[t]
\centering 
\includegraphics[width=0.40\textwidth]{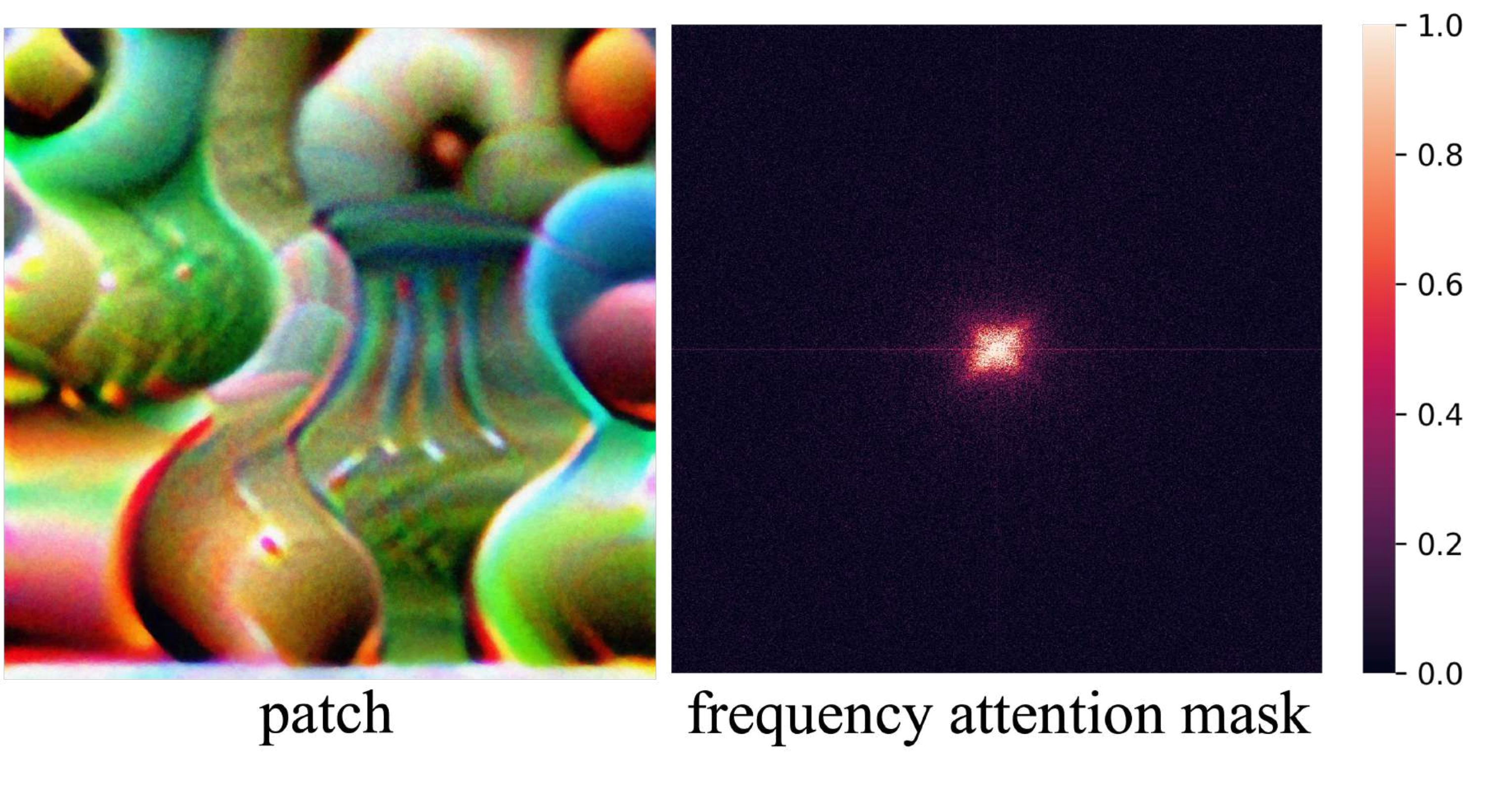} 
\caption{Frequency attention mask of the patch.} 
\label{fig.frequency} 
\end{figure}

In this paper, we hope that the patch can successfully attack objects of various sizes, which means that it must remain adversarial even after being shrunk in several times. 
The most commonly used image reduction algorithm is bilinear interpolation.
Because this algorithm is a Low-pass filter, patch's high-frequency information will be lost during the reduction process.
As a result, patch attack capability is dwindling. 
We used FRAN to guide the patch generation process, and the experimental results demonstrate the workability of the method.
ASRm and ASRs can achieve 96.41\% and 87.21\% for small objects using the $mathcal{P}_2$ attack the detector.
Compared to the baseline, the attack success rate with $\mathcal{P}_1$ is only 92.52\% and 83.03\%. 
For small objects, ASRl of $\mathcal{P}_1$ and $\mathcal{P}_2$ is 96.75\% and 96.98\%, respectively. 
We improved the attack success rate of small and medium objects while not affecting large objects' attack success rate. 
Figure \ref{fig.frequency} depicts the spectrum attention map when attack YOLOv3. 
Because Cb and Cr channels in YCbCr color space represent color offset, we only show Y channel's frequency attention mask.

\subsubsection{Attack against different object detectors}

In this section we compared the result on three object detectors: YOLOv3\cite{IEEEexample:Redmon2018YOLOv3}, YOLOv4\cite{IEEEexample:YOLOv4} and RetinaNet\cite{IEEEexample:RetinaNet}.
For all these detectors, we used the same dataset and the same training process as described in Section \ref{section.set}. The results were displayed in the table 2.

\begin{table*}[h]
	\centering
	\caption{The ASR (\%) of adversarial patches generated from Ours, Ours(non-YCbCr), Ours(keep) and baseline\cite{IEEEexample:Thys2019FoolingAS} in digital-world against YOLOv3. FRAN is the option of whether to use the frequency domain attention mask, YCbCr represents whether to convert the patch into YCbCr color space, Keep use FRAN indicates whether the frequency attention mask is used to guide the generation of patches along all the training process.}
	\label{table.abtest}
	\begin{tabular}{ccclrrrr}
	\toprule
	method          & FRAN                  & YCbCr                & \multicolumn{1}{c}{Keep use FRAN} & \multicolumn{1}{c}{ASR(\%)} & \multicolumn{1}{c}{ASRl(\%)} & \multicolumn{1}{c}{ASRm(\%)} & \multicolumn{1}{c}{ASRs(\%)}  \\ 
	\hline
	Ours            & $\checkmark$         & $\checkmark$         &                                  & \textbf{93.68}              & 96.75                        & \textbf{96.41}               & \textbf{87.21}                \\
	Ours(non-YCbCr) & $\checkmark$         &                      &   & 85.33                        & 89.27                        & 92.07                        & 78.93                         \\
	Ours(keep)      & $\checkmark$         & $\checkmark$         & \multicolumn{1}{c}{$\checkmark$} & 88.02                       & 94.29                        & 90.88                        & 77.70                         \\
	Baseline        & \multicolumn{1}{l}{} & \multicolumn{1}{l}{} &                                  & 91.12                       & \textbf{96.98}                        & 92.52                        & 83.03                         \\
	Random          & \multicolumn{1}{l}{} & \multicolumn{1}{l}{} &                                  & 21.76                       & 23.62                        & 16.60                        & 25.01                         \\
	Clean           & \multicolumn{1}{l}{} & \multicolumn{1}{l}{} &                                  & 17.83                       & 19.26                        & 13.61                        & 20.69                         \\
	\bottomrule
	\end{tabular}
\end{table*}

\begin{table}[htbp]
	\centering
	\label{table2}
	\caption{The ASR(\%) against different object detectors. Attacking three object detectors separately and comparing the effect of the two methods (Ours, Baseline\cite{IEEEexample:Thys2019FoolingAS})}
	\begin{tabular}{crrrr} 
	\toprule
	Detector            & \multicolumn{1}{c}{ASR(\%)} & \multicolumn{1}{c}{ASRl(\%)} & \multicolumn{1}{c}{ASRm(\%)} & \multicolumn{1}{c}{ASRs(\%)}  \\ 
	\hline
	YOLOv3(Ours)        & 93.68                       & 96.75                        & 96.41                        & 87.21                         \\
	YOLOv3(Baseline)    & 91.18                       & 96.98                        & 92.52                        & 83.03                         \\
	RetinaNet(Ours)     & 84.31                       & 90.37                        & 85.94                        & 75.55                         \\
	RetinaNet(Baseline) & 81.78                       & 90.11                        & 80.85                        & 72.99                         \\
	YOLOv4(Ours)        & 63.57                       & 70.43                        & 62.15                        & 57.12                         \\
	YOLOv4(Baseline)    & 61.44                       & 70.53                        & 56.14                        & 53.34                         \\
	\bottomrule
	\end{tabular}
\end{table}

\subsection{Ablation Study}

This section performed additional experiments to better understand our method's robustness. 
In the first set of experiments (clean), the images were used without a patch. 
In the second set of experiments (random), the images were used with a random noise patch.
The third set of experiments (baseline) employed the baseline\cite{IEEEexample:Thys2019FoolingAS} method.
The fourth set of experiments (Ours(nonYCbCr)) utilizes FRAN without converting it to YCbCr color space.
The fifth set of experiments (Ours(keep)) used FRAN throughout the generating process.
The sixth set of experiments (Ours) only utilized FRAN in the first seven epochs.
The results of ablation experiments were depicted in table \ref{table.abtest}.

\subsubsection{Robustness against JPEG compression}

According to Ian et al., JPEG compression\cite{IEEEexample:Goodfellow2015ExplainingAH}, which removes the image's high-frequency signals, can significantly reduce the perturbability of attacks.
The adversarial image's effectiveness can be lowered by JPEG compression, which removes the image's high-frequency signals.

To identify how JPEG compression affects adversarial patch attacks, we saved the patch in JPEG format and reloaded it to calculate their attack success rates on the test dataset. 
The results were displayed in the table \ref{t.jpg}.

\begin{table}[h]
\begin{center}
\caption{The impact of JPEG compression on attacks}
\label{t.jpg}
\begin{tabular}{lrrrr} 
\toprule
Methods      & \multicolumn{1}{l}{ASR(\%)} & \multicolumn{1}{l}{ASRl(\%)} & \multicolumn{1}{l}{ASRm(\%)} & \multicolumn{1}{l}{ASRs(\%)}  \\ 
\hline
Ours                &  \textbf{93.68}  &   96.75   &  \textbf{96.41}   &  \textbf{87.21}     \\
Ours(JPEG)          &  ($\downarrow$5.65)\textbf{88.03}  &  ($\downarrow$2.45)\textbf{94.30}    &   ($\downarrow$5.53)\textbf{90.88}   &  ($\downarrow$9.51)\textbf{77.70}     \\
Baseline\cite{IEEEexample:Thys2019FoolingAS} &  91.18       &  \textbf{96.98}     &  92.52    & 83.03          \\
Baseline\cite{IEEEexample:Thys2019FoolingAS}(JPEG) & ($\downarrow$7.63)83.55  &  ($\downarrow$4.03)92.95  &  ($\downarrow$7.55)84.97 & ($\downarrow$11.81)71.12                          \\
\bottomrule
\end{tabular}
\end{center}
\end{table}

The experiment results reveal that our method outperforms the baseline when attacking small and medium-sized objects. 
The attack success rate can be increased by directing patch creation with FRAN.
However, the attack success rate decreases if FRAN is used throughout the training process.
The reason is that FRAN suppressed frequency information excessively during the later training period, reducing patch aggression. 
So we chose to use FRAN duing the first several epochs.
When attacking different detectors we will choose diiferent epoch to stop using FRAN.
At the same time, converting the patch into YCbCr color space and extracting frequency information using Fourier transform during the training phase result in a patch with a greater attack impact.

\section{Conclusion and Future Work}   

\label{Conclusion and Future Work}

A frequency attention module is proposed in this work to guide the patch generating process.
When attacking the YOLOv3 object detector, our attack approach has a success rate of 93.68\%. 
The attack success rate of medium and small targets has improved without compromising the attack success rate on big targets. 
Our findings show that frequency attention can control the frequency distribution when creating patches, resulting in more robust patches.

In the future we would like to extend this work by making it more powerful.
One way to do this is optimizing the generating process of frequency attention map.(i.e. use a generate model to produce the frequency attention map)
Another way is to optimize the training process of frequency attention module.
Our current method will suppress too much frequency information, using drnamic weight might improve upon this.

\bibliographystyle{unsrt}  
\bibliography{references}

\end{document}